\documentclass[letterpaper, 10 pt, conference]{ieeeconf}  

\IEEEoverridecommandlockouts                              

\usepackage{graphics} 
\usepackage{epsfig} 
\usepackage{mathptmx} 
\usepackage{times} 
\usepackage{amsmath} 
\usepackage{amssymb}  
\usepackage{hyperref}  

\usepackage{flushend}
\usepackage{mars}

\title{\LARGE \bf
Heterogeneous Multi-sensor Calibration based on Graph Optimization
}

\author{Hongyu Chen$^{1}$ and S\"oren  Schwertfeger$^{1}$
\thanks{$^{1}$Authors are with School of Information Science Technology of ShanghaiTech University
{\tt\small <chenhy3, soerensch>@shanghaitech.edu.cn}}%
}

\begin{document}

%
%


\marsPublishedIn{Accepted for:} 		

\marsVenue{IEEE International Conference on Real-time Computing and Robotics(RCAR) 2019}

\marsYear{2019}

\marsPlainAutors{ Hongyu Chen, S\"oren Schwertfeger}


\marsMakeCitation{Heterogeneous Multi-sensor Calibration Based on Graph Optimization}{IEEE Press}


\marsIEEE{}


\makeMARStitle

\maketitle
\thispagestyle{empty}
\pagestyle{empty}

\begin{abstract}
Many robotics and mapping systems contain multiple sensors to perceive the environment. Extrinsic parameter calibration,  the identification of the position and rotation transform between the frames of the different sensors, is critical to fuse data from different sensors. When obtaining multiple camera to camera, lidar to camera and lidar to lidar calibration results, inconsistencies are likely.  We propose a graph-based method to refine the relative poses of the different sensors. We demonstrate our approach using our mapping robot platform, which features twelve sensors that are to be calibrated. The experimental results confirm that the proposed algorithm yields great performance. 
\end{abstract}

\section{INTRODUCTION}

Nowadays,  robot platforms are typically equipped with multiple sensors,  such as multiple cameras, 3D lidar sensors and IMUs, which require calibration in order to present sensed information in a common coordinate system.  Cameras provide rich color and feature information, while lidar sensors can provide accurate distances.  Cameras only work in appropriate lighting conditions and problems may happen at night.  Lidars are active sensors, they emit infrared light and sense the reflections to detect objects in
the environment, thus they can work in the dark.  For robot tasks such as Simultaneous Localization and Mapping (SLAM) \cite{cadena2016past}, it is important to know the relative pose of each sensor to the robot base and to each other.  


Many approaches to calibrate different types of sensors have been proposed.  

Calibration methods can be divided into online and offline calibration.   Online calibration means that the relative poses of the sensors can be computed during the usage of the system.  However, offline methods can usually provide more accurate extrinsic parameters.  

The camera to camera calibration, it can be divided into two groups. One group requires an overlap between the two cameras. It is a well studied problem \cite{heikkila1997a, tsai1987a, Zhang2000A}.  Zhang \cite{Zhang2000A} proposed a method to estimate intrinsic and extrinsic parameters at the same time.   There is a well-known Matlab calibration toolbox named Caltech camera calibration toolbox. 

Another group is that of non-overlapping cameras. Kumar \cite{Kumar2008Simple} utilizes mirrors to create virtual views of the calibration board.  Such a method is flexible, however, it requires elaborate configuration.  Carrera \cite{carrera2011slambased} utilizes an external SLAM system to compute extrinsic parameters. SLAM system based algorithms typically are not very accurate, because of wrong feature matching, especially in environments with repetitive structures.  This will decrease the accuracy of the camera pose estimation. 

Both lidar and cameras can provide rich and complementary data, which can be used in various tasks.  For the transformation between 3D lidar and cameras, Dhall \cite{Dhall2017Lidar} utilizes a special target to compute extrinsic parameters.  Correct 3D-3D  corresponding points from the camera frame and the lidar frame are needed. Pandey \cite{pandey2010extrinsic} utilizes a planar checkerboard pattern to compute the extrinsic parameters, with the assumption that the normal vector of the plane coincides with the normal vector of the 3D point cloud on the plane. They use these constraints to form a non-linear optimization problem, which was optimized by using the Levenberg Marquardt (LM) \cite{Levenberg1944A, marquardt1963an} algorithm. 

Martin \cite{calibrationrgbcamerawithvelodyneLidar} utilizes one planar board that has four circular holes in front of a blackboard.  The holes in both the acquired image data and the 3D Lidar point cloud were detected automatically. This is flexible if the point cloud acquired from the 3D lidar is dense. However, we cannot apply this method to the point cloud measured by some 3D lidar sensors if the acquired 3D point is sparse. Ishikawa \cite{ishikawa2018lidar} proposes a method of target-less and automatic camera lidar calibration.  Camera motion is estimated by feature detection and matching while lidar motion is calculated using the ICP \cite{BeslA} algorithm.  The motion of 3D lidar may not be very accurate, since ICP algorithm is easy to fall into local optimum.

For multi-sensor calibration, Heng \cite{heng2013camodocal, heng2014infrastructurebased} proposes an online multi-camera extrinsic calibration  method using  multiple view geometry and bundle adjustment.  Sungdae \cite{sim2014closed} is using closed loop constraints for multi-sensor calibration.

In order to calibrate 3D lidar sensors and cameras, we exploit Ranjith's \cite{fast} approach.   Similar to Pandey \cite{pandey2010extrinsic}, we also need one planar target.  The two-stage estimation process is utilized to compute the translation part and the rotation part. In the second stage, a non-linear function is formed to jointly optimize the translation part and rotation part. For the calibration of two 3D lidars, we adopt a Normal Distribution Transformation {NDT} \cite{Biber2003The} algorithm to compute the extrinsic parameters with an initial guess. 

Hand-Eye calibration is a well know calibration problem. The goal of the conventional Hand-Eye calibration problem is to compute the transformation between a gripper (hand) and a camera (eye)\cite{tsai1987a}. One solution to this problem is to use bundle adjustment on identified feature points in the camera image and on the to-be-optimized transformation between the robot arm and the camera simultaneously \cite{zhi2017simultaneous}. 

Using a fully calibrated sensor system it is possible to employ applications such as sensor fusion \cite{pathak20073d} and SLAM \cite{cadena2016past}, especially in three dimensions \cite{birk20093}.

The  contributions of this paper  are as follows:
\begin{itemize}
  \item We present a novel approach to calibrate multiple types of sensors based on graph optimization, with a focus on the accuracy of the extrinsic calibration. 
  \item We describe an extend hand-eye calibration method for non-overlapping cameras, especially also for the calibration of the camera w.r.t. our external, camera-based tracking system.
  \item We present synthetic and real-data experiments to show the performance of our algorithm. Our mapping robot has 12 different sensors, which are calibrated well using our approach.
\end{itemize}

The remainder of the paper is organized as follows: Section \ref{sec:problem} is giving the problem description. Section \ref{sec:method} describes our algorithm while Section \ref{sec:experiments} presents the simulation and real robot experiments. The conclusions are drawn in Section \ref{sec:conclusions}.

\section{Problem Description}
\label{sec:problem}

An essential assumption for most calibration approaches is, that all sensors are rigidly mounted on the robot platform. Extrinsic calibration then means the estimation of the relative poses of sensor pairs, such that all the data collected from the different sensors can be fused into one single frame.  Due to sensor noise, it is impossible to align the data without error. Also, when using real sensors, there is typically no way to accurately measure the transform (translation and rotation) of the physical sensor to another frame. This is because the sensor frame is typically somewhere inside the sensor, and because there are no tools available to measure arbitrary translations and rotations of physical objects with sufficient accuracy. 

\begin{figure}[tb]
	\centering 
	\includegraphics[width=1\linewidth]{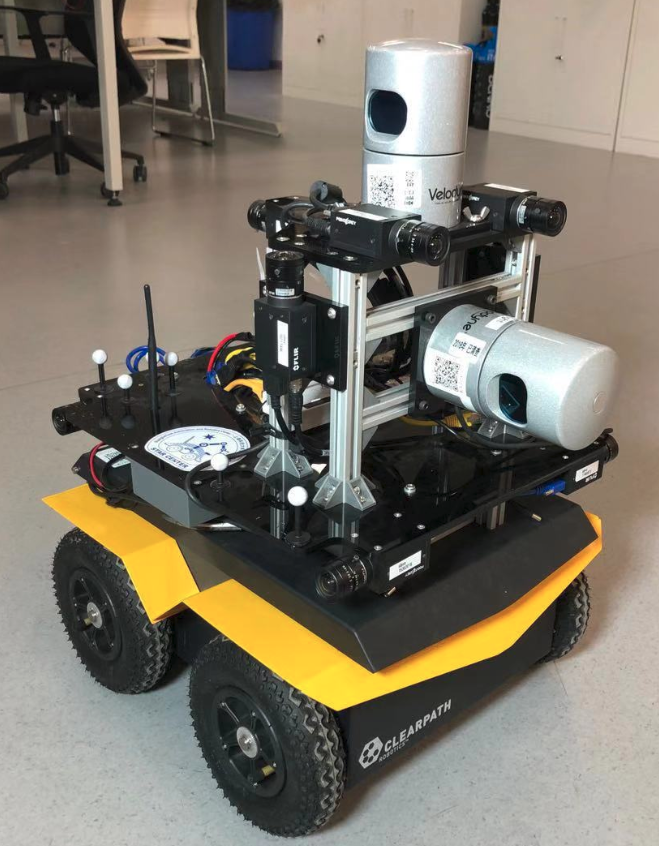}  
		\caption{The MARS Mapper robot with its sensors, that is calibrated in this paper.   }
	\label{fig:robot_platform}   
\end{figure}

The sensor platform employed in this paper, the MARS Mapper (MARS is the acronym of our Mobile Autonomous Robotic Systems Lab), is shown in Figure \ref{fig:robot_platform}. The MARS mapper features of two Velodyne HDL-32E lidar, nine FLIR Grasshopper3 cameras with 5 megapixel resolution, and one Xsens MTi-100 IMU. Among these twelve sensors,  one of the lidar scans the horizontal plane and the other lidar scans the vertical plane. The cameras are mounted in pairs for stereo vision to the front, the left, the right and up. Additionally, there is a monocular camera facing backwards.

All cameras and the IMU are hardware synchronized using a custom micro-processor. This device also provides exact time-stamps to the Velodyne lidars, by simulating GPS time-stamps. Additionally, the system is also able to be hardware synchronized with the tracking system. Using a single Intel i7 CPU, the robot is able to gather and compress (JPEG Quality 90) the 5 MP image data of the 9 cameras with 10Hz, synchronized with the data-collection of the two lidars, the IMU and odometry data.


\section{Method}
\label{sec:method}

We first describe the pairwise calibration methods below and then the global multi-sensor calibration in Section \ref{sec:global}.

\subsection{Pairwise Calibration}
\label{sec:pairwise}

The pairwise calibration is used to estimate the $SE(3)$ transformation between sensor pairs. 

The following calibration pairs are considered and described in more detail below (the number of pairs for our real robot experiment is shown in parentheses):
\begin{itemize}
 \item Stereo-camera (overlapping) (4)
 \item Non-overlapping cameras (32) 
 \item 3D lidar to camera (13)
 \item 3D lidar to 3D lidar (1)
 \item Tracking system to camera (9)
\end{itemize}

\subsubsection*{Stereo-camera calibration}
Stereo cameras have, by definition, a mostly overlapping field of view. There are many stereo calibration algorithms known in literature \cite{strecha2008benchmarking}. We are employing the algorithm from \cite{Zhang2000A} in our system.

\subsubsection*{Non-overlapping camera pairs}
We apply Hand-Eye calibration to calibrate non-overlapping cameras.  
Suppose we have a series of corresponding motions of the hand and the eye, and $A_{ij}$ and $B_{ij}$ are the homogeneous transformation of the eye and hand motions, respectively.  Let $X$ be the transformation from hand to eye.  We need at least two motions to formulate the Hand-Eye equation \cite{tsai1987a}. 
\begin{equation}
\label{formula1}
A_{ij}X = XB_{ij}
\end{equation}

As each homogeneous transform has the following form 

\begin{equation}       
\left(               
\begin{array}{cc}  

R & t\\  
0 & 1\\  

\end{array} 
\right)  \in SE(3)              
\end{equation}

where $R \in SO(3)$ and $t \in $$\mathbb{R}^3$ denote the rotation and translation, respectively. 

We can split Eq.~\ref{formula1} into a rotation and a translation part.

\begin{equation}
\label{formula2}
R_aR_x=R_xR_b
\end{equation}
\begin{equation}
\label{formula3}
R_at_x+t_a =R_bt_b +t_x
\end{equation}

Many researchers propose different methods to solve Eq. \ref{formula2} and \ref{formula3}, 
such as \cite{daniilidis1999handeye}, \cite{tsai1987a}, \cite{horaud1995haye}.



We extend the Hand-Eye calibration to calibrate two non-overlapping cameras, since our mapping robot contains stereo cameras and also non-overlapping cameras. With two cameras rigidly mounted, we assume that one camera is the eye and the other camera is the hand.  Having calibrated the intrinsic parameters beforehand, we can now estimate the motion of the cameras by using two checkerboards, one for each camera.  Once we get a series of camera motions, the transformation of the two non-overlapping cameras can be estimated.


\subsection*{3D Lidar to Camera Calibration}

The goal of our 3D lidar to camera calibration is to find the extrinsic parameters from the laser range coordinate frame to the camera frame. We assume the usual pin-hole camera model. The relationship between a homogeneous point $P_w = [x_w, y_w, z_w, 1]$  in the world coordinates  and its coordinate image position $p_c = [u, v, 1]$ is given by:

\begin{equation}
p_c = K(R_cP_w + t_c)
\end{equation}   
where  $R_c$ and $t_c$ are the extrinsic parameters related to the world coordinate frame and the camera coordinate frame.  $K$ is the camera intrinsic matrix, $R_c$ a $3$x$3$ orthonormal matrix representing the camera's orientation.  $t_c$ is a $3$ vector representing the camera's position.   

The laser range finder returns the distance measurements to the points on the planner board we use for calibration. Suppose  $P_l$ in the laser range finder coordinate frame is located at $P_c$ in the camera coordinate frame.   The transformation from the laser range finder coordinate frame to the camera frame can then be described by:

\begin{equation}
P_c = R_lP_l + t_l
\end{equation}
where $R_l \in SO(3)$ and
$t_l \in\mathbb{R}^3$ 
denote the rotation and translation, respectively.

For 3D lidar to camera calibration, we explore Hebert's \cite{unnikrishnan2005fast} calibration method. At first, the translation and rotation part are considered separately. In the second stage, a non-liner function is used to refine the translation and rotation part.

\subsection*{3D Lidar to 3D Lidar}

Calibration between two 3D lidars can be considered as a registration problem for two 3D point clouds.   With an initial guess, the NDT algorithm is applied to compute the relative pose of the two 3D lidars.  


To check the calibration result of two 3D lidar, we adopted one planner board and calculate the normal of two 3D lidars in the 
 frames named $n_1$ and $n_2$, respectively.  For the rotation part, suppose that $T$ is the transformation between the 3D lidars.  For that we use the following formula:
\begin{equation}
 \label{estimate2}
 D = \frac{n_1*n_2}{||n_1||||n_2 ||}
\end{equation} 
Here $D$ must be close to $1$ if the rotation part of the calibration result is accurate. The closer to 1 D is, the more parallel the
two normal vectors are, which is positively correlated with the rotation
part of the calibration result.  For the translation part, we utilize the following error function $E_l$ to verify the calibration result:
 \begin{equation}
 \label{estimate1}
E_l = \frac{1}{N }  \sum_{i=1}^{N}( p_{t}^{i} - R_p * p_{s}^{i} -t_p )
 \end{equation}
where $R_p \in SO(3)$  and $t_p$  denote the rotation and
translation, respectively. $p_{t}$ and $p_{s}$ are points from two lidars.
The smaller $E_l$ is, the better the result.

\subsubsection*{Tracking System Sensors}

To gather additional constraints for the later optimization we also collect calibration data using the tracking system. For this we attach several markers on the robot platform, which form a frame of reference rigid to the robot body. Now Hand-Eye calibration is applied to estimate the transformation from this tracking system frame on the robot to each of the sensors. The motion of the markers, and thus the frame of reference on the robot, can be precisely estimated by the tracking system, while the motions of cameras are estimated by the checkerboards.

\subsection{Global Calibration}
\label{sec:global}

Many existing approaches try to compute the transformation of a pair of sensors.  Suppose we need to calibrate three sensors.  The sensor $A$ and sensor $B$ pair is calibrated first.  Repeatedly, we can calibrate the sensor $B$ and sensor $C$ pair. The transformation between sensor $A$ and sensor $C$ can be calculated by chaining the two previous calibrations, or it can be directly calibrated.  In each calibration process, there is an error, which can be propagated to the next step.  

The extrinsic parameters calculated by combining the two sensor pair transformations will differ from the transformation obtained from a direct calibration. A global calibration is needed to reduce the total errors. 
In the previous sections, we have computed the relative poses of each sensor pair. In the following we describe how to achieve a global calibration using those results.

We are building a hypergraph to do the global calibration, which contains several nodes and edges. The hypergraph used for calibrating multiple sensors is shown in Fig. \ref{fig:posge_graph}. 
 A spanning tree of the pairwise calibration results is used as the initial guess.
In the big graph, we assume that all the sensors poses are unknown.  

In this work, our hypergraph $G(V, E)$ contains several nodes $V$ and hyperedges $E$.  A node $v \in V | v \in SE(3)$ contains the sensor pose while an edge $e \in E$ denotes the relative transformation between two (sensor) frames.   The goal of the graph optimization is to find the poses of $V$ that minimize all the errors over all edges which can fit all the available measurements.  Define that node $x_i \in V$ provides the poses where $i$  is the node identity. The hyper edge $e_i \in E$  denotes the relation between the sensors.  Then we can form the error function we want to minimize in the following form: 

\begin{figure}[tb]
	\centering 
	\includegraphics[width=0.5\linewidth]{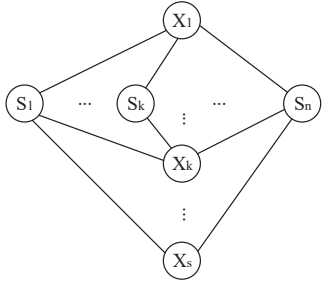}  
	\caption{
		The graph representing the relationships between different sensors.  $X_1 ...  X_s$ and $ S_1 ...  S_n $ are different types of sensors such as camera and Velodyne.  An edge between two nodes represents a direct sensor-to-sensor calibration between these two devices.}
	\label{fig:posge_graph} 
\end{figure}

\begin{equation}
F(x) = \sum_{v_i,v_j\in V} e(x_i, x_j, u_{ij})^T \Omega_{ij} e(x_i, x_j, u_{ij})
\end{equation}

\begin{equation}
\label{formula}
\hat{x} = \mathop {\arg\min }\limits_x  F(x)
\end{equation}

where  $u_{ij}$ is the initial constraint of node $i$ and $j$. $\Omega_{ij}$ represents the information matrix of the constraint. 

The graph optimization is aimed to minimize Eq.~\ref{formula}.  During optimization the Jacobian of the error function $F(x) $ will be computed. Then a linear step will be taken to find the poses of sensors which minimize the error of the function. 
With an initial guess $\hat{x}$, the solution of \ref{formula} can be found by iteratively  solving a linear system with the system matrix $H$ and the vector $b$, such that 
\begin{equation}
H = \sum_{i, j \in V} J_{ij(\hat{x})}^T \Omega_{ij} J_{ij(\hat{x})}
\end{equation}
\begin{equation}
b^T = \sum_{i, j \in V} e_{ij}^T \Omega_{ij} J_{ij(\hat{x})}
\end{equation}. 
Here $J$ is the Jacobian of the error function, with an initial guess $\hat{x}$. To effectively solve the non-linear function we use G2O \cite{kummerle2011g}.


\subsection{Uncertainty of the Transform between each Pair}

The uncertainty  of transformation between each pair of sensors  comes  from  two main  sources. $(i)$ The uncertainty of the method we used to find the transformation in the pair-wise sensor calibration.  $(ii)$  The uncertainty of the data we collected from different sensors.  In Hand-Eye calibration, the pose estimation of hand and eye have some uncertainties. Also, the method we used to compute the transformation contains some uncertainties. For different types of sensor pairs,  the calibration algorithm can give us the best transformation between two sensor pairs. For convenience, we combine the transformation between one sensor pair into one vector $\rho=[\delta_x, \delta_y, \delta_z, \phi_x, \phi_y, \phi_z]$. 

The Jackknife resampling method \cite{Quenouille1949Approximate} is used to calculate the variance of $\rho$.
This method is performed by taking samples from the entire data pairs we collected to calibrate the two sensors. We use 3D lidar to camera calibration as an example.  Suppose $N_i$ is the $i_{th}$ 3D lidar image pair, and $m$ is the total number of 3D lidar and image pairs we used for calibration. We omit the $i_{th}$ sample  $X_i = [N_1, . . . , N_{i-1}, N_{i+1}, . . . , N_m]$ from the data for the Jackknife method. For each of these samples $X_i$, a different parameter vector  $\rho_i$ is obtained. The parameter variance is given by:
\begin{equation}
\sigma_p^2 = \frac{m-1}{m} \sum_{i=1}^{m}(\rho_i - \widehat{\rho})
\end{equation} 
 
where $\widehat{\rho}= \sum_i p_i/n$, $p$ is the transformation between sensor pairs.  The inverse of the variance matrix is the 
information matrix we used in our global optimization. The information matrix  is $\Omega\in \Lambda^6$,   which is a diagonal matrix.

\begin{figure}[tb]
	\centering 
	\includegraphics[width=0.8\linewidth]{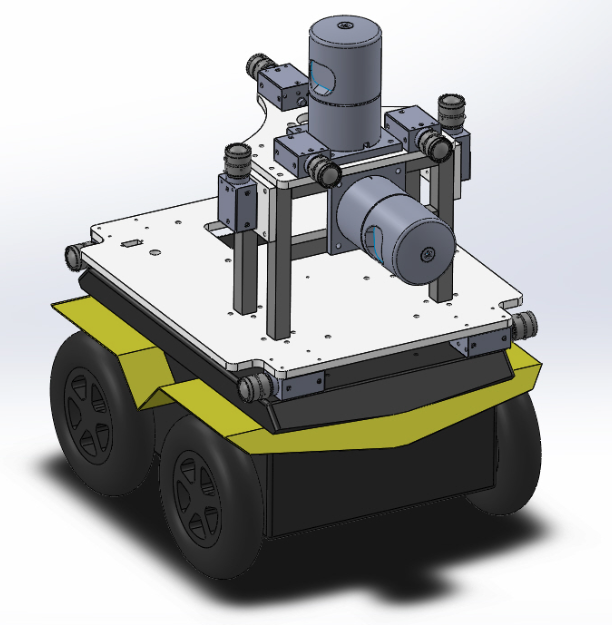}  

	\caption{CAD model of MARS Mapper robot. }
	\label{fig:before_fused_points} 
\end{figure}

\section{Experimental Results }
\label{sec:experiments}

We evaluate our method on both synthetic data and real data. For the simulation tests we  know the relative pose between the sensors. We implement our algorithm on G2O \cite{kummerle2011g}, a framework for graph optimization. For the real data tests, our mapping robot platform as shown in Fig. \ref{fig:robot_platform} is used to test our method. Bouguet's camera calibration toolbox \cite{sgfd} is utilized to obtain the intrinsic parameters of the cameras.

\subsection{Experiments with Synthetic Data}
The following experiments are designed to show that the proposed algorithm is able to work efficiently on multi-sensor calibration. 
We first evaluate our approach with synthetic data. A hyper-graph is build with $4$ vertices and $6$ edges. For all the edges between vertices, a zero-mean Gaussian noise $N(0,\sum)$ is added to the transformation.

\begin{figure}
	\begin{minipage}[t]{1.0\linewidth}
		\centering
		\includegraphics[width=1.0\linewidth]{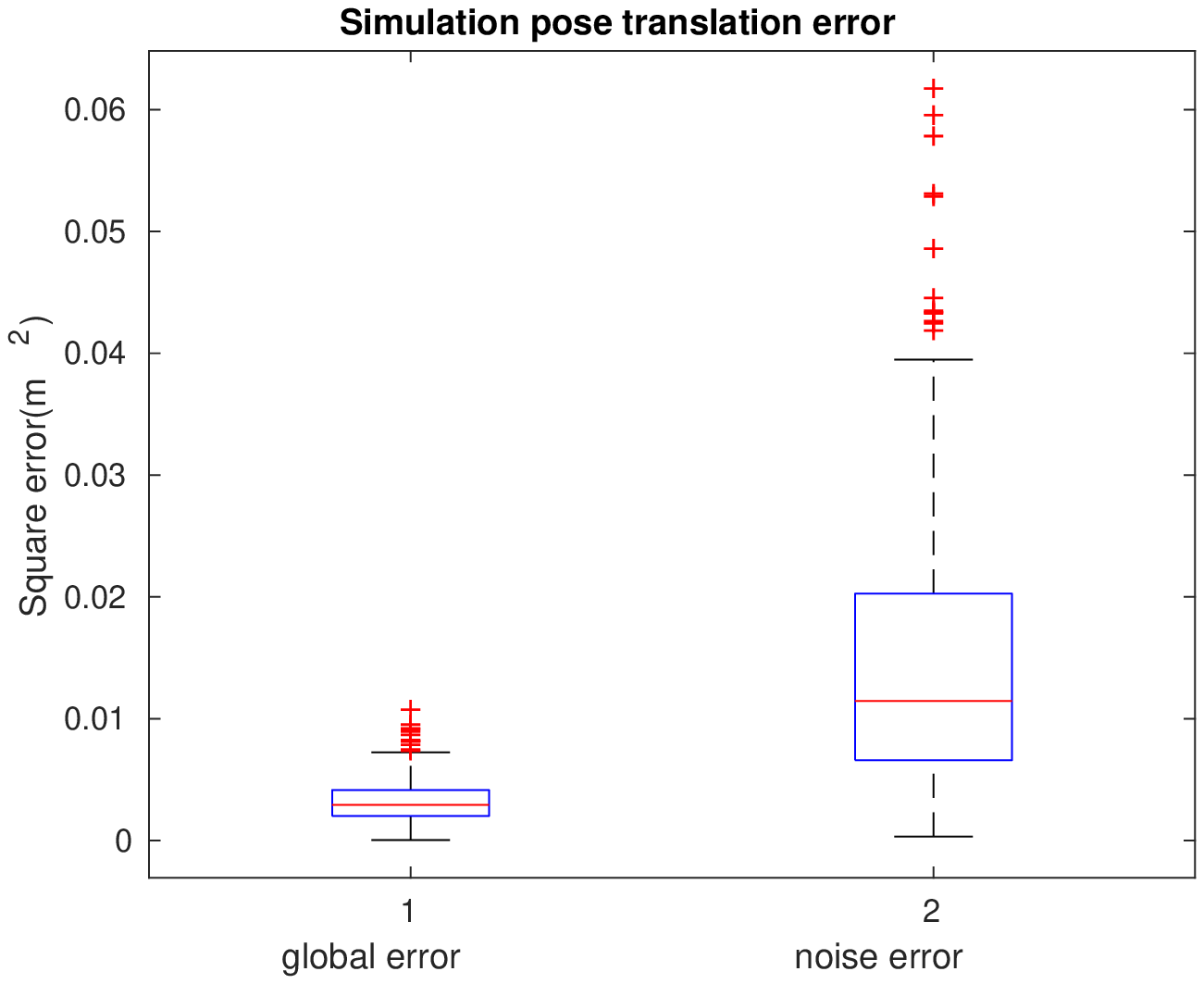}  
		\caption{(a) Optimized translation error and noise error added to edges.  }
		\label{fig:trans}
	\end{minipage}%
	\vspace{.05in} 
	\begin{minipage}[t]{1.0\linewidth}
		\centering
		\includegraphics[width=1.0\linewidth]{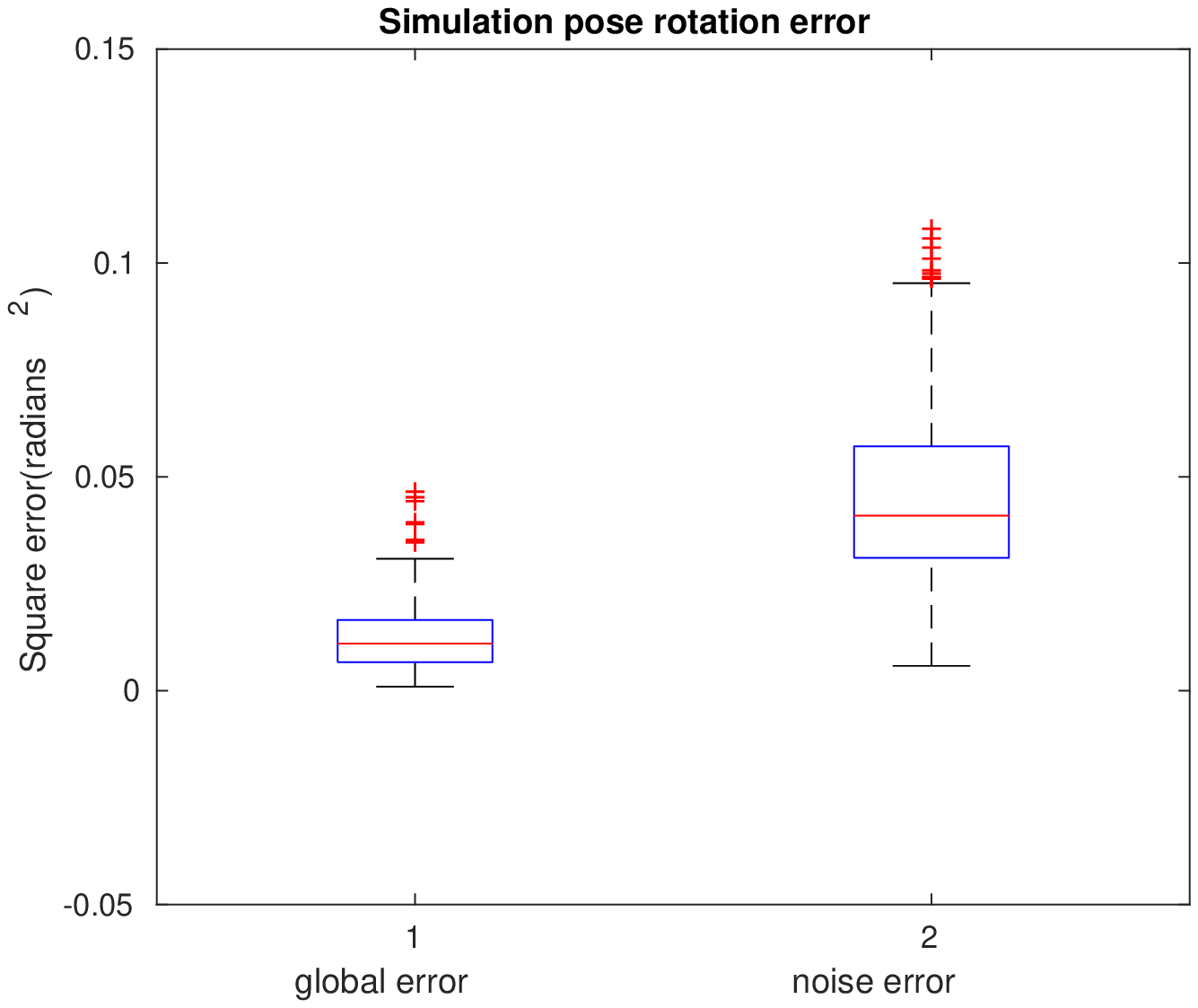}  
		
	\end{minipage}
	\caption{Optimized rotation error and noise error added to edges. }
	\label{fig:rot}
\end{figure}

We generate $300$ datasets to evaluate the performance of our approach. The error of the graph is calculated by the sum of the squared errors between  all edges. The rotation part between all edges and the translation part are considered separately. The error of the translation part is presented in terms of meters, so the error unit we shown in Fig.~\ref{fig:trans} is $m^2$. For the rotation part we adopt angle-axis to represent the transformation. 
Fig.~\ref{fig:trans} and Fig.~\ref{fig:rot} show the translation error and rotation error after the global optimization result on the left. Additionally we see the noise that was added to the edges, which simulate the pair-wise calibration results. We can see that the variance of the global optimization is lower than the noise we add to sensor transformations.

\subsection{Experiments with Real Data}

\begin{figure}
		\centering
		\includegraphics[width=1.0\linewidth]{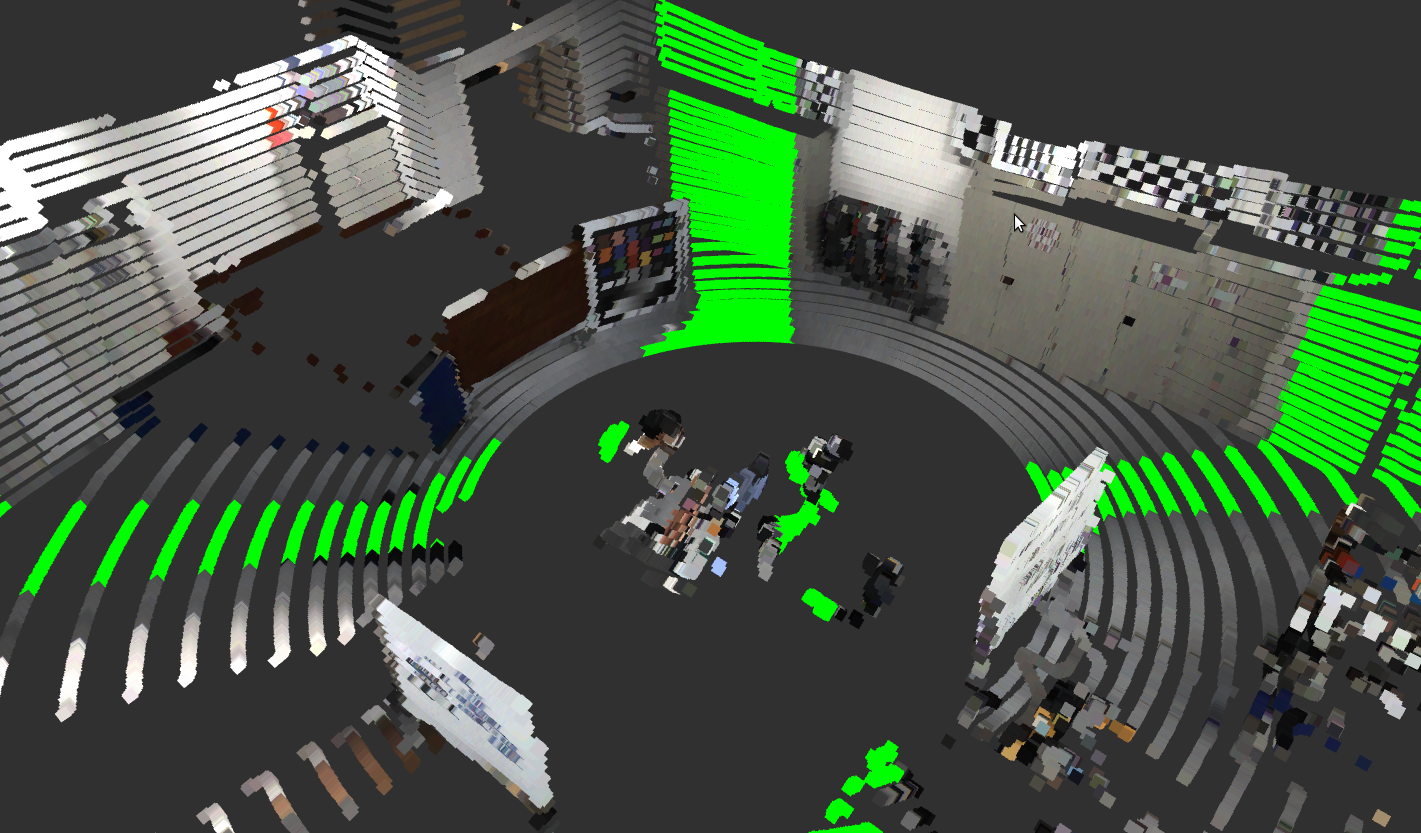}  
		
	\caption{ A 3D Lidar scan colored by all the $7$ horizontal cameras. The transformations between each camera and 3D lidar are acquired from the global optimization result. All the green points represent areas where no camera is overlapping with the point cloud. }
   \label{fig:coloredpoints}
\end{figure}

\begin{figure}
	\begin{minipage}[t]{0.5\linewidth}
		\centering
		\includegraphics[width=1.6in]{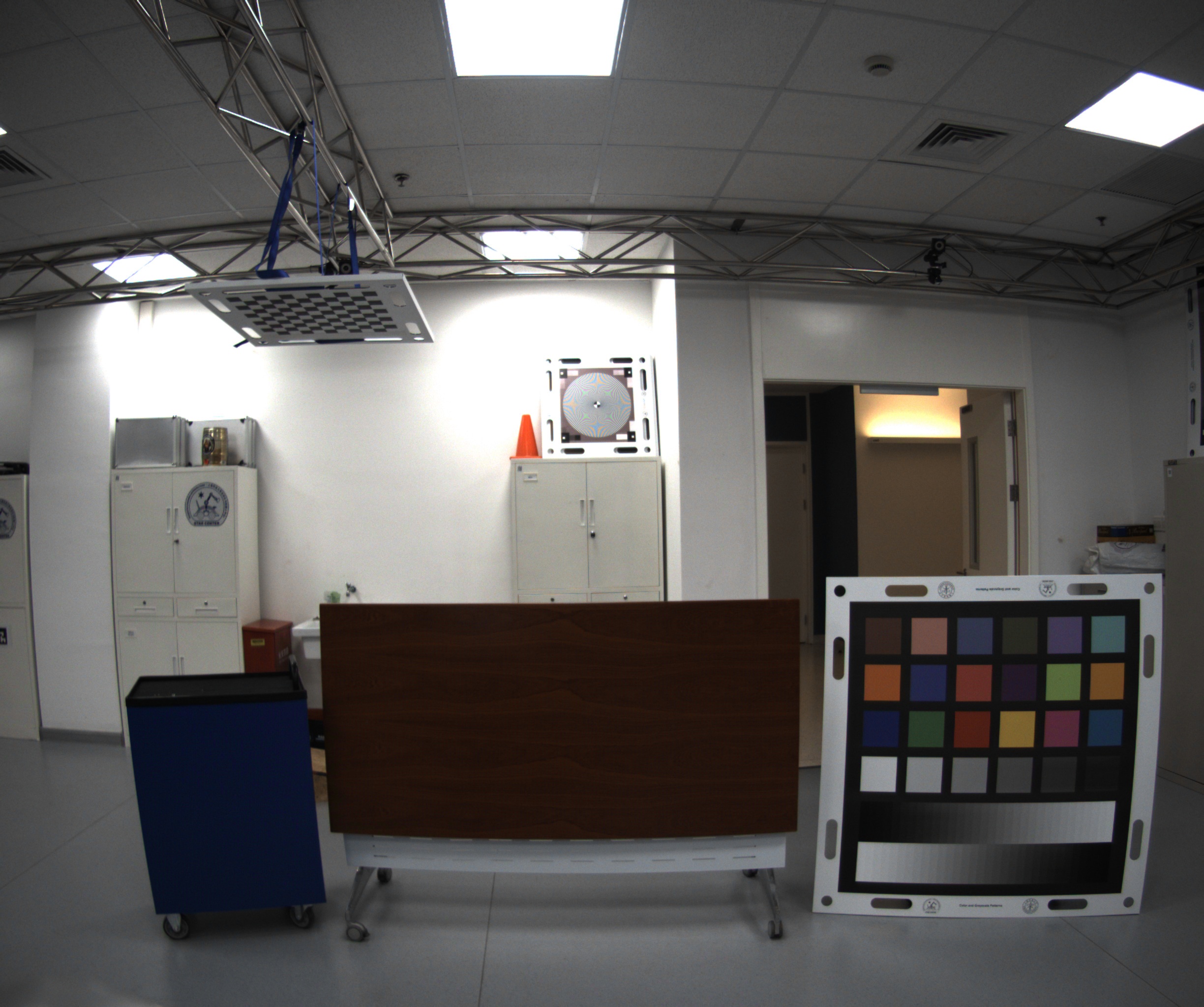}
	\end{minipage}%
	\begin{minipage}[t]{0.5\linewidth}
		\centering
		\includegraphics[width=1.6in]{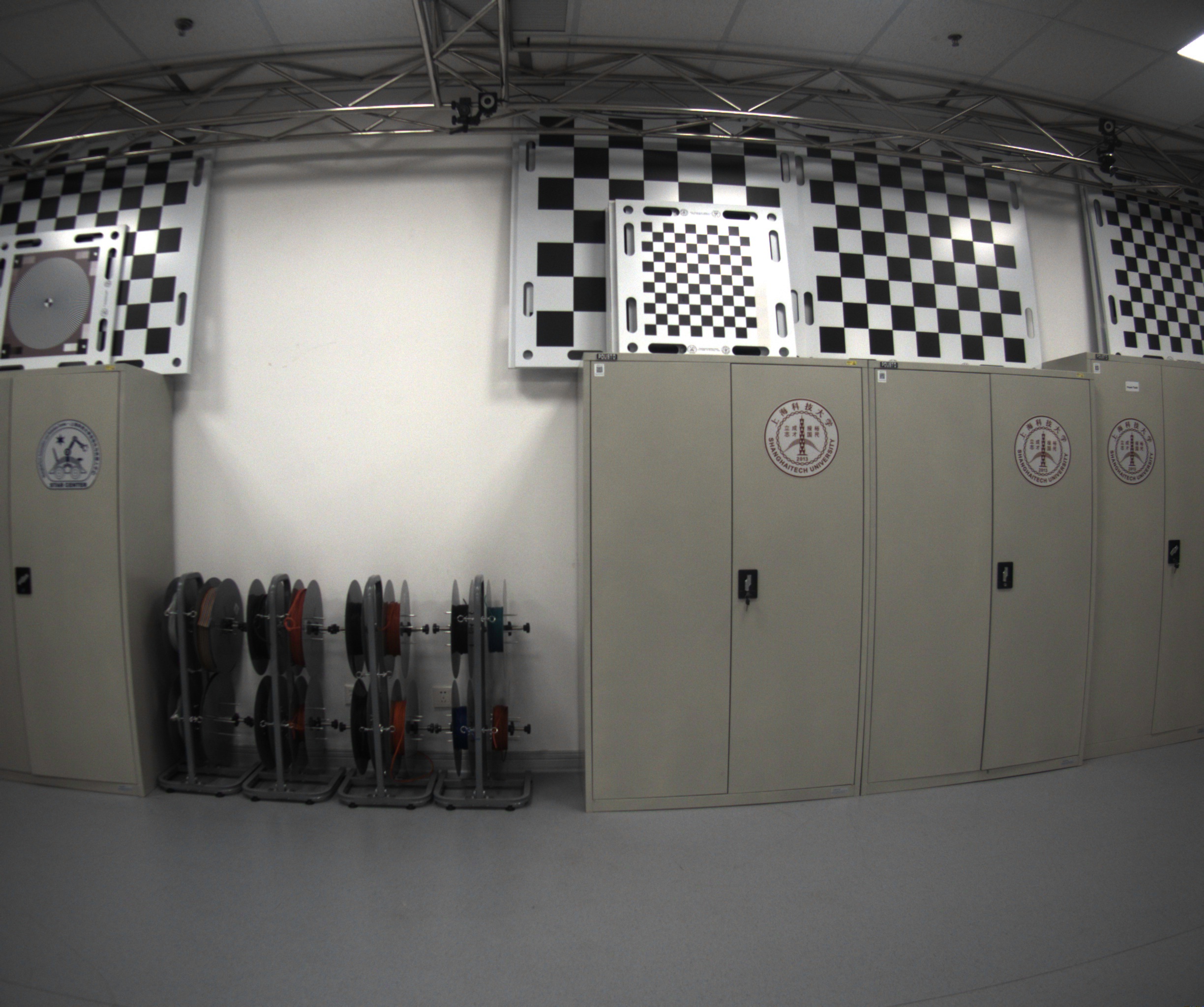}

	\end{minipage}
	\caption{Image acquired from two cameras. The left image is acquired from one of the left side cameras while the  right image is acquired from one of the front cameras.}
   \label{fig:normalimage}
\end{figure}

With the 9 cameras mounted on the MARS Mapper system, as shown in Fig.~\ref{fig:robot_platform}, there are 36 pairs of cameras. Among these 36 camera-pairs, four of those pairs are calibrated using stereo calibration. The two cameras in each stereo camera pair share a common field of view, which are calibrated with the algorithm proposed in \cite{Zhang2000A}. 
In addition, the other 32 pairs of non-overlapping cameras are calibrated using the Hand-Eye calibration described above. We collected camera data for these pairs with different poses of our mapping robot platform. 

Besides, we also used our Optitrack tracking system \footnote{\url{https://www.optitrack.com/products/prime-13/}} with 21 Prime 13 cameras fixed to a truss on the ceiling to calibrate the 9 cameras on the robot. 
For 3D lidar to camera calibration, we can only calibrate lidar to camera pairs that share a common field of view. Those are $6$ cameras with the horizontal 3D lidar and $7$ cameras for the vertical 3D lidar.

With the robot moving around, we can estimate the accuracy of two 3D lidar calibration results by using formulas \ref{estimate2} and \ref{estimate1}. Once all the sensor pairs are calibrated, a hypergraph is built with $12$ vertices and $59$ edges. G2O is used to optimize the graph. In order to evaluate the optimized calibration result, the points acquired from the vertical 3D lidar are fused with the $7$ cameras that overlap its field of view. 
We propose an global optimization approach to optimization all the sensors. The following formula \ref{error} is used to calculate the error.
 
\begin{equation}
 e_{ij} = ln( T_{ij}^{-1}  T_{i}^{-1} T_{j})^{\vee}
\end{equation} 
where $\vee$ is Lie Algebra operation $T_{ij}$ is the transformation between sensor $i$ and sensor $j$,$ T_{i}$ and $T_{j}$ are initial pose for sensor $i$ and sensor $j$. 
\begin{equation}
\label{error}
E_{global} = \frac{1}{N_E} \sum_{i = 1}^{N_E}e_{i}^2 
\end{equation}
where $E_{global}$ is the global error. $N_E$ is the total number of edges in our graph, $e_{i}$ is the error for each edge. 
 Fig. \ref{fig:coloredpoints} shows the colored point cloud  that was generated by fusing the range data of the lidar with the color data of the cameras, using the global, optimized calibration result.
 Table \ref{tab:mytable} shows that the error decreased after optimization.
 
\begin{table}[]
	\caption{ Error before optimization and after optimization}
	\label{tab:mytable}
	\begin{tabular}{lllll}
		\cline{1-3}
		\multicolumn{1}{|l|}{}      & \multicolumn{1}{l|}{before optimization} & \multicolumn{1}{l|}{after optimization} &  &  \\ \cline{1-3}
		\multicolumn{1}{|l|}{error} & \multicolumn{1}{l|}{0.0149}              & \multicolumn{1}{l|}{0.0126}             &  &  \\ \cline{1-3}                                          
	\end{tabular}
\end{table}
 
  Fig. \ref{fig:normalimage} shows two example images from the data we collected with the cameras. In  Fig. \ref{fig:normalimage}, the left image contains a color checker board. In Fig. \ref{fig:coloredpoints} we can see the very well represented color board in the colored point cloud. Also, other parts of the point cloud look well calibrated. Green points in the point cloud indicate points which are not covered by any camera of the mapping robot.



		



\section{Conclusions}
\label{sec:conclusions}
In this paper, we presented a graph based algorithm for multi-sensor calibration of a mapping robot platform with many sensors. The proposed algorithm requires known sensor to sensor calibration before graph optimization. We estimate the covariance matrices of the hyper-edges of the graph using the Jacknife resampling algorithm. Experiments on synthetic data show quantitatively, that the optimization improves the calibration. Our real robot experiment demonstrates the good performance of our algorithm. In the future, we plan to add more sensor types such as IMU, 2D lidar and also the motion base of our robot to our hypergraph.

\bibliographystyle{IEEEtran}
\bibliography{ref.bib}

\end{document}